\documentclass[sigconf]{acmart}
\AtBeginDocument{%
  \providecommand\BibTeX{{%
    Bib\TeX}}}
\setcopyright{acmlicensed}
\copyrightyear{2025}
\acmYear{2025}
\acmDOI{XXXXXXX.XXXXXXX}
\acmConference[Conference acronym 'XX]{Make sure to enter the correct
  conference title from your rights confirmation email}{June 03--05,
  2025}{Woodstock, NY}
\acmISBN{978-1-4503-XXXX-X/2018/06}

\begin{document}

\title{ParlAI Vote: A Web Platform for Analyzing Gender and Political Bias in Large Language Models}

\author{Wenjie Lin}
\authornote{Equal contribution.}

\affiliation{%
  \institution{Purdue University}
  \city{}
  \country{}
}
\email{lin1790@purdue.edu}

\author{Hange Liu}
\authornotemark[1]

\affiliation{Independent Researcher%
  \institution{}
  \city{}
  \country{}
}
\email{gary.liuhg@gmail.com}

\author{Yingying Zhuang}

\affiliation{%
  \institution{Johns Hopkins University}
  \city{}
  \country{}
}
\email{yzhuang6@jh.edu}

\author{Xutao Mao}
\affiliation{%
  \institution{Vanderbilt University}
  \city{}
  \state{}
  \country{}
}
\email{xutao.mao@vanderbilt.edu}

\author{Jingwei Shi}
\affiliation{Independent Researcher
 \institution{}
 \city{}
 \state{}
 \country{}}
 \email{shijingwei0712@gmail.com
}

\author{Xudong Han}
\affiliation{%
  \institution{MBZUAI}
  \city{}
  \state{}
  \country{}}
  \email{xudong.han@mbzuai.ac.ae}

\author{Tianyu Shi}
\affiliation{%
  \institution{University of Toronto}
  \city{}
  \state{}
  \country{}}
\email{ty.shi@mail.utoronto.ca}

\author{Jinrui Yang}
\authornote{Corresponding author.}
\affiliation{%
  \institution{The University of Melbourne}
  \city{}
  \country{}}
\email{jinryang@unimelb.edu.au}



\renewcommand{\shortauthors}{Lin et al.}

\begin{abstract}
We present ParlAI Vote \footnote{Demo: \url{https://euro-parl-vote-demo.vercel.app/} \\ Video: \url{https://www.youtube.com/@Jinrui-sf2jg}}, an interactive web platform for exploring European Parliament debates and votes, and for testing LLMs on vote prediction and bias analysis. This web system connects debate topics, speeches, and roll-call outcomes, and includes rich demographic data such as gender, age, country, and political group. Users can browse debates, inspect linked speeches, compare real voting outcomes with predictions from frontier LLMs, and view error breakdowns by demographic group. Visualizing the EuroParlVote benchmark and its core tasks of gender classification and vote prediction, ParlAI Vote highlights systematic performance bias in state-of-the-art LLMs. It unifies data, models, and visual analytics in a single interface, lowering the barrier for reproducing findings, auditing behavior, and running counterfactual scenarios. This web platform also shows model reasoning, helping users see why errors occur and what cues the models rely on. It supports research, education, and public engagement with legislative decision-making, while making clear both the strengths and the limitations of current LLMs in political analysis.
\end{abstract}





\maketitle

\section{Introduction}
LLMs are now widely used to read and analyze political text~\cite{potter2024hidden, li2024political}, such as text classification~\cite{liu2024poliprompt}, stance detection~\cite{lan2024stance,wagner2024power}, coalition negotiation modeling~\cite{moghimifar2024modelling}, and debate analysis~\cite{liu2024llm,heide2025understanding}.
These uses lead to two core needs. First, they need practical tools to examine whether model performance is valid and equitable across demographic groups.~\citet{von2024united} shows that LLMs fail to reliably predict individual voting behavior in the 2024 European Parliament elections, with accuracy varying across contexts and requiring detailed prompts, highlighting strong limitations and biases in using LLM-synthetic samples for public opinion prediction.~\citet{rettenberger2025assessing} also find that open-source LLMs show consistent political bias on European Union issues: larger models align more with left-leaning parties while smaller models are closer to neutral. Second, researchers require a transparent way to connect language to votes rather than treating prediction as a black box~\cite{liang2024cmat,wang2024enhancing,yi2025score,zhou2024human,zhou2025reagent}.

The European Parliament is a transparent and practical setting for such work because debates, roll-call votes, and member attributes are public and detailed~\cite{van2016debates}, but these signals are spread across sources and formats, which raises the cost of systematic analysis and teaching. There is a developed website, HowTheyVote~\cite{howtheyvote}, where users can find summaries and votes of Members of the European Parliament (MEPs). To further evaluate LLMs in politically sensitive contexts, we previously introduced the EuroParlVote Benchmark~\cite{yang2025benchmarkinggenderpoliticalbias}, which systematically links debate topics and speeches to roll-call outcomes and includes rich demographics.

Until now, however, there has been no research-oriented and AI-powered visualization system that links debate content to legislative outcomes and demographics while also supporting model-based evaluation. On top of this dataset, we build an interactive web platform, ParlAI Vote, that enables exploration of issue-speech-vote connections. The system integrates frontier and open-source LLMs to simulate MEPs' voting behavior. This combination of data visualization and predictive AI creates a novel environment for investigating political behaviors. This system allows researchers and even the general public to intuitively understand how debate-related contents relate to voting outcomes, while demonstrating the potential and limitations of LLMs in political science. Our system makes three contributions:
\begin{itemize}
    \item We introduce ParlAI Vote, which, to the best of our knowledge, is the first AI-powered and unified web system that unifies debates, roll-call votes, and demographic data of the European Parliament into a live, explorable interface, making a complex political benchmark accessible to researchers and the public.
    

    \item Our demo lets users experiment with different LLMs of their choice to predict MEPs' voting behavior from debates, enabling comparative exploration of model capabilities and biases while highlighting both the potential and risks of AI in democratic decision-making.
    \item Our system offers an intuitive way for users to observe how LLMs implicitly encode stereotypical associations and biases. By turning gender and political bias evaluations into interactive exploration, the demo makes issues of fairness tangible, fostering transparency and accountability discussions in AI for societal domains.
\end{itemize}

\section{Web Platform Description}

To manage the dataset of 969 roll-call votes, the main homepage interface provides essential search, filter, and sorting tools. These features are crucial for allowing users to quickly isolate specific legislative events or analyze political trends over time. It also works as a navigation hub, giving users an overview of debates across years and topics, and allowing smooth movement from broad exploration to focused inspection of a single vote.

The vote page provides an integrated interface for exploring roll-call votes in the European Parliament. At the top, users see the debate title, metadata (vote date, report ID, number of participants), and the overall outcome. The centerpiece is the Vote Breakdown visualization, which displays how MEPs voted. By default, votes are grouped by political affiliation and ordered along the left–right spectrum, making ideological alignments immediately visible. Users can pivot this breakdown by country, gender, or age, enabling analysis of how demographics intersect with political behavior and highlighting subgroup divergences that are invisible in aggregate results.

\begin{figure}[ht]
    \centering
    \includegraphics[width=0.7\linewidth, angle=270]{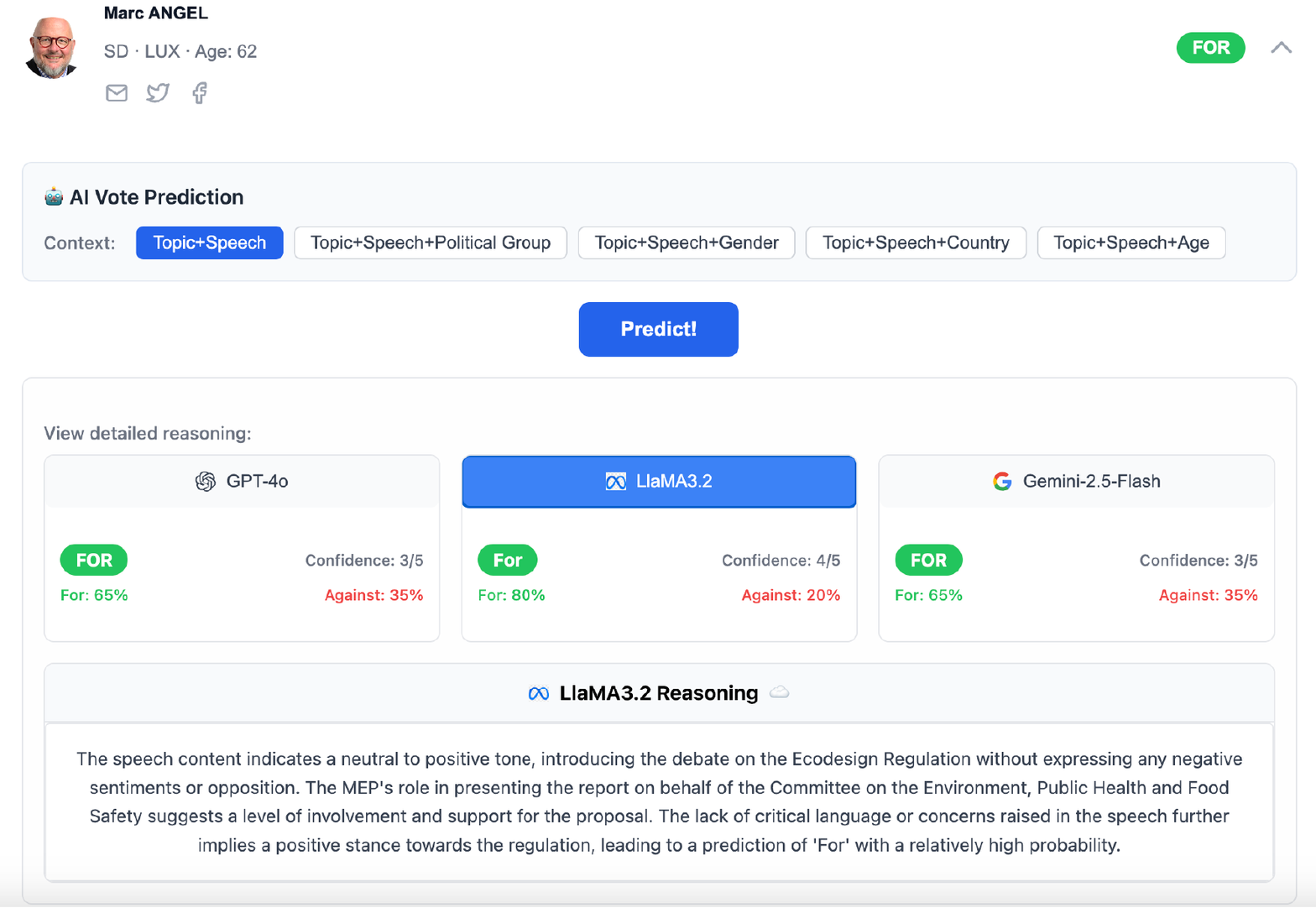}
    \caption{Vote Prediction results on Ecodesign Regulation by Marc Angel when providing Topic and Speech referred as context and using Llama-3.2 as LLM predictor.}
    \label{fig:predict}
\end{figure}

The core of the system is the set of interactive AI Prediction modules, which include both Vote Prediction and Gender Prediction. For each debate speaker (Figure~\ref{fig:predict}), users can:

\begin{enumerate}
\item \textbf{Access Ground Truth:} Display the MEP’s actual vote and/or gender as a baseline reference. This anchoring in real-world outcomes is essential for assessing model reliability, making it clear not only if a model is wrong but also how its errors are distributed across different groups.

\item \textbf{Demographic Impact Explorer:} Predict votes or gender using only the debate speech, then compare outcomes when demographic attributes are introduced. A counterfactual mode allows users to modify variables (e.g., adding political group or demographic information), making fairness concerns visible by showing how sensitive model predictions are to individual attributes.

\item \textbf{Compare Models:} Run multiple LLMs side-by-side on the same input to examine differences in accuracy, robustness, and bias. All models were run with the same settings, using a temperature of 0.3 and a maximum output length of 512 tokens. This direct comparison helps users understand how proprietary and open-weight models behave under identical conditions. The system keeps the interaction consistent across models so that differences reflect model behavior rather than interface differences.

\item \textbf{Inspect Reasoning:} Read the model-generated reasoning for each prediction to understand why a decision was made. This helps reveal whether models rely on meaningful argumentation or superficial cues. The reasoning view supports both qualitative and quantitative analysis of model decision patterns and reduces the opacity of model predictions.
\end{enumerate}

\section{Findings}\label{Findings}
Leveraging this system~\cite{yang2025benchmarkinggenderpoliticalbias}, we find consistent evidence of gender bias: female MEPs are disproportionately misclassified as male in gender prediction, and vote simulation performance drops significantly when speakers are labeled as female. Proprietary models such as GPT-4o and Gemini-2.5 demonstrate lower misclassification rates and more balanced treatment of female speakers, whereas open-weight models (e.g., Llama-3.2, Mistral) exhibit stronger male bias and weaker ~\cite{yang2025benchmarkinggenderpoliticalbias}. Attempts to mitigate these disparities through LoRA \cite{hu2021lora} fine-tuning were unsuccessful, often worsening performance for female MEPs.

On the political dimension, we observe that LLMs display a centrist bias. Models simulate voting behavior of centrist and liberal groups most accurately, while struggling with ideologically extreme groups. 
Interestingly, far-right groups are predicted more reliably than far-left groups, suggesting that rhetorical uniformity in right-aligned discourse may be easier for models to capture.
Providing explicit political group identifiers improves fairness by boosting accuracy on underrepresented and extreme groups. Collectively, these findings highlight the persistent demographic and ideological biases of LLMs and underscore the importance of context-aware, multilingual benchmarks for assessing fairness in politically sensitive applications \cite{feng2023pretraining, rozado2024political, santurkar2024opinions}.

\section{Error Analysis and Causes of Bias}

To better understand why LLMs produce systematic gender and political errors in ParlAI Vote, we extend our analysis beyond accuracy and examine the reasoning generated by the models. Because the web platform records the models’ explanations for every prediction, it allows detailed inspection of how specific mistakes arise. We focus on error cases from GPT-4o, which is the strongest model in earlier evaluations. From 792 cases where the model predicted with high confidence (confidence score $\geq$ 4/5) but produced an error, we conducted both quantitative and qualitative analyses of the reasoning content. This reveals stable patterns that help explain the sources of bias observed in Section~\ref{Findings}.

\subsection{Causes of Gender Bias}

\begin{table}[t]
\centering
\caption{Stereotypical linguistic cues observed in the model’s reasoning among 792 high-confidence errors in gender prediction\protect\footnotemark. Note that one error may contain multiple stereotypical terms.}

\label{gender_language}
\resizebox{\columnwidth}{!}{
\begin{tabular}{lcc}
\toprule
\textbf{Stereotypical Term} & \textbf{Assumed Gender} & \textbf{Occurrences in Errors} \\
\midrule
assertive         & Male   & 515 \\
direct            & Male   & 372 \\
structured        & Male   & 268 \\
confrontational   & Male   & 209 \\
technical         & Male   & 209 \\
emotional         & Female & 183 \\
personal          & Female & 156 \\
empathetic        & Female & 148 \\
\bottomrule
\end{tabular}
}
\end{table}

\footnotetext{We acknowledge gender is non-binary, but use a male/female classification here, as it is an accurate representation of past and present MEPs according to the metadata.}

A major source of gender errors comes from how models attach gender labels to specific linguistic cues. The frequencies of these errors are reported in Table~\ref{gender_language}. In many incorrect cases, the model infers gender from tone, sentiment, and language style rather than content. Terms such as \textit{assertive}, \textit{direct}, \textit{structured}, \textit{confrontational}, or \textit{technical} lead the model to confidently assign a male prediction, while words like \textit{emotional}, \textit{personal}, or \textit{empathetic} lead to female. These patterns echo well-documented linguistic stereotypes, including work by Deborah Tannen~\cite{tannen1990you}, where report-style language is seen as male-coded and rapport-style language as female-coded. The model appears to follow a similar rule. When a female MEP uses firm or technical language, the model treats these cues as strong signals of masculinity even when the true label is female. 

In addition to language style, topic stereotypes also contribute to errors. The model sometimes assumes that subject matter reveals gender. As shown in Table~\ref{gender_topic}, policy areas such as economic regulation, migration, or geopolitical issues are often mapped to male, while topics such as human rights or women’s rights are mapped to female. This produces systematic mistakes: a female MEP speaking on economic policy is often labeled as male, and a male MEP discussing human rights is sometimes labeled as female. This topic–gender mapping is incorrect and indicates that the model has absorbed patterns from pre-training that reinforce existing stereotypes.

\begin{table}[t]
\centering
\caption{Topic keywords that the model associates with male- or female-coded assumptions during gender prediction errors. Frequencies indicate how often each topic appears in reasoning traces that lead to misclassification.}
\label{gender_topic}
\resizebox{\columnwidth}{!}{
\begin{tabular}{lcccc}
\toprule
\textbf{Topic Keyword} & \textbf{Stereotype Gender } & \textbf{Male Pred.} & \textbf{Female Pred.} & \textbf{Total} \\
\midrule
economic            & Male    & 145 & 31 & 176 \\
geopolitical        & Male    & 63  & 5  & 68  \\
human rights        & Female  & 3   & 65 & 68  \\
women's rights      & Female  & 0   & 33 & 33  \\
gender-mainstreaming & Male   & 8   & 0  & 8   \\
migration policy    & Male    & 2   & 0  & 2   \\
\bottomrule
\end{tabular}
}
\end{table}


\subsection{Causes of Political Bias}

We also examine errors for political groups, focusing on the ``European Conservatives and Reformists'' (ECR) and ``Identity and Democracy'' (ID) groups (right and far-right), which show the largest drops in accuracy~\cite{yang2025benchmarkinggenderpoliticalbias}.
By inspecting the explanations generated by GPT-4o, Llama-3.2, and Gemini-2.5, we identify three recurring error categories.

\begin{figure}[t]
    \centering
    \includegraphics[width=\columnwidth]{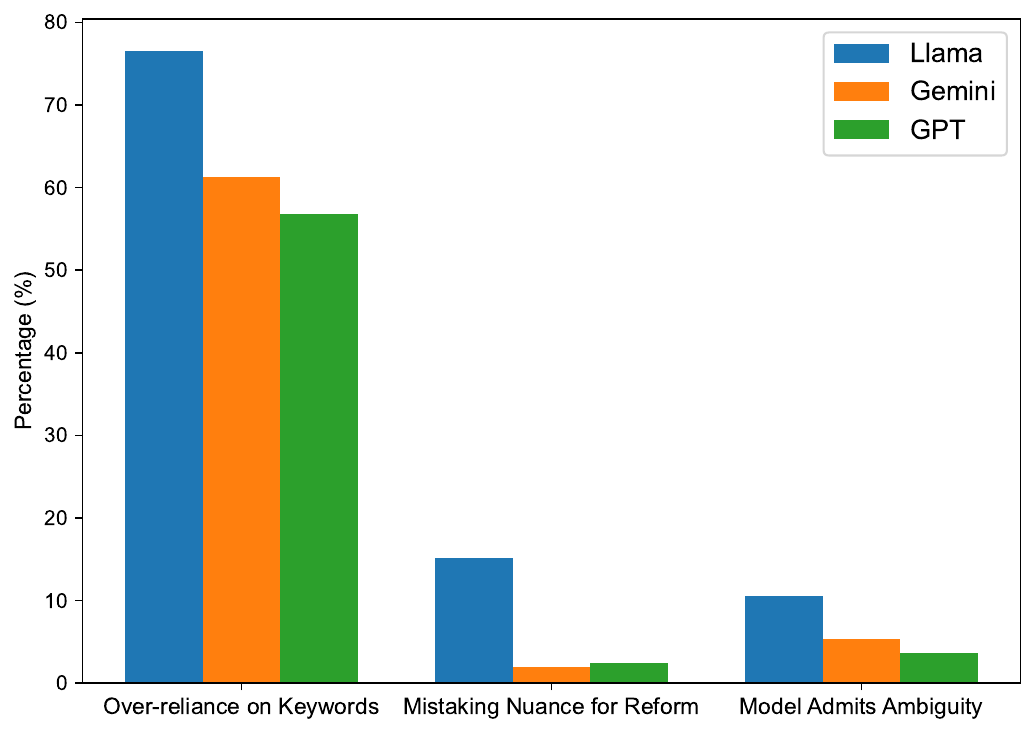}
    \caption{Failure categories across LLMs. The x-axis lists the three error types, and the y-axis shows the percentage of errors that fall into each category. Note the sum is not equal to 100\% since there are still other error reasons.}
    \label{bar_failure_categories}
\end{figure}

The errors for political groups come from three connected patterns in how models process the speeches. First, models rely heavily on specific keywords and treat them as direct signals for a vote, mapping terms such as \textit{national sovereignty} or \textit{protecting borders} to an \textit{against} prediction and words like \textit{climate} or \textit{human rights} to \textit{for} prediction, which fails when these terms are used to support the opposite position. Second, models often interpret ideological criticism as reform-oriented support, because centrist and left groups use criticism to argue for improvements, while right-leaning groups may use similar criticism to reject a proposal; as a result, when an ID or ECR member describes a proposal as \textit{bureaucratic} or \textit{damaging to member state authority}, models predict \textit{for} even though the intention is to oppose it. Third, when models are unsure, they tend to make a \textit{for} prediction, which reflects a reliance on broad language cues rather than specific political contexts of debates. This default choice creates additional errors for right-aligned groups, whose rhetorical patterns do not match the models' simple templates.

As shown in Figure~\ref{bar_failure_categories}, these failure modes appear at different frequencies across the models. Specifically, keyword mistakes are the most common category, followed by misinterpreting criticism as desire for reform, and then a smaller fraction of cases where the model openly states uncertainty but still defaults to \textit{for} prediction. Some error cases include more than one of these factors.

\section{Conclusions}

ParlAI Vote visualizes and extends the EuroParlVote Benchmark into an end-to-end workflow for investigating how parliamentary language connects to vote outcomes. The web system enables users to trace the full chain of evidence: from a debate topic, navigating to linked speeches, and examining the corresponding roll-call votes. Predictions from LLMs can be placed side-by-side with the ground truth, allowing users to compare outcomes and uncover systematic discrepancies. Additionally, the interface surfaces error distributions across demographic groups, exposing hidden fairness gaps. It further provides counterfactual probes that let users test ``what-if'' scenarios and model rationales that clarify decision pathways, creating opportunities for deeper interpretability. In conclusion, these features of this web make it easier to evaluate, diagnose, and assess model behavior in political contexts. The extended analysis shows why these errors occur by connecting the model reasoning to linguistic cues, topic stereotypes, and vote prediction heuristics. By linking system-level interaction with detailed reasoning analysis, the web interface reveals not only where LLMs fail but also how these failures arise, offering a clearer path for future improvement.

\section*{Ethic Statement}


This demo is a research tool designed to analyze bias in multilingual LLMs using open-sourced European Parliament data. It does not make factual or actionable claims about individuals and includes clear disclaimers to prevent misuse for opinion influence or demographic manipulation. Gender is treated as metadata, and the counterfactual mode is used only for bias inspection. The dataset contains only public legislative records and speeches and does not include private or sensitive personal information, so no informed-consent procedures are required. This work involves no human subjects and follows all ACM data-use policies.

\bibliographystyle{ACM-Reference-Format}
\bibliography{sample-base}

\end{document}